\documentclass[sigconf]{acmart}

\usepackage[font=small,labelfont=bf]{caption}
\usepackage{hyperref}
\usepackage{tikz}
\usepackage{amsmath}
\usepackage{filecontents}
\usepackage{footnote}
\usepackage{diagbox}
\usepackage{listings}
\usepackage{booktabs}
\usepackage{multirow}
\usepackage[ruled,vlined]{algorithm2e}
\usepackage{subfigure}
\usepackage{tabularx}
\usepackage{url}
\usepackage{algorithmic}
\usepackage{graphicx}
\usepackage{textcomp}
\usepackage{xcolor}
\usepackage{verbatim}
\usepackage{enumitem}
\usepackage{pifont}
\usepackage{listings}
\usepackage[framemethod=tikz]{mdframed}
\usepackage{balance}
\usepackage{lipsum} 
\usepackage{indentfirst}
\usepackage[htt]{hyphenat}
\usepackage{tcolorbox}

\AtBeginEnvironment{tcolorbox}{\small}

\usepackage{siunitx}
\def\etal{\emph{et al.}\xspace}

\AtBeginDocument{%
  \providecommand\BibTeX{{%
    \normalfont B\kern-0.5em{\scshape i\kern-0.25em b}\kern-0.8em\TeX}}}

\copyrightyear{2023} 
\acmYear{2023} 
\setcopyright{acmlicensed}\acmConference[WWW '23 Companion]{Companion Proceedings of the ACM Web Conference 2023}{April 30-May 4, 2023}{Austin, TX, USA}
\acmBooktitle{Companion Proceedings of the ACM Web Conference 2023 (WWW '23 Companion), April 30-May 4, 2023, Austin, TX, USA}
\acmPrice{15.00}
\acmDOI{10.1145/3543873.3587321}
\acmISBN{978-1-4503-9419-2/23/04}

\begin{document}

\title{Copyright Protection and Accountability of Generative AI: \\Attack, Watermarking and Attribution}

\author{Haonan Zhong$^{1}$$^{\mathsection}$, Jiamin Chang$^{1}$$^{\mathsection}$, Ziyue Yang$^{2}$$^\mathsection$, Tingmin Wu$^{3}$, \and  Pathum Chamikara Mahawaga Arachchige$^{3}$, Chehara Pathmabandu$^{3}$, and 
Minhui Xue$^{3}$} 

\affiliation{
\institution{$^{1}$University of New South Wales, Australia}
\institution{$^{2}$The Australian National University,  Australia}
\institution{$^{3}$CSIRO's Data61, Australia}
\country{}
}

\renewcommand \authors{Haonan Zhong, Jiamin Chang, Ziyue Yang, Tingmin Wu,  Pathum Chamikara Mahawaga Arachchige, Chehara Pathmabandu, and Minhui Xue}

\renewcommand{\shortauthors}{Zhong et al.}

\begin{abstract}
   %
    Generative AI (e.g., Generative Adversarial Networks – GANs) has become increasingly popular in recent years. However, Generative AI introduces significant concerns regarding the protection of Intellectual Property Rights (IPR) (resp. model accountability) pertaining to images (resp. toxic images) and models (resp. poisoned models) generated. In this paper, we propose an evaluation framework to provide a comprehensive overview of the current state of the copyright protection measures for GANs, evaluate their performance across a diverse range of GAN architectures, and identify the factors that affect their performance and future research directions. Our findings indicate that the current IPR protection methods for input images, model watermarking, and attribution networks are largely satisfactory for a wide range of GANs. We highlight that further attention must be directed towards protecting training sets, as the current approaches fail to provide robust IPR protection and provenance tracing on training sets.
    
\end{abstract}


\maketitle 

\def\thefootnote{$^{\mathsection}$}\footnotetext{The first three authors made equal contributions, and the work was done at CSIRO's Data61, Australia.}\def\thefootnote{\arabic{footnote}}
\pagestyle{plain}

\section{Introduction}

%
The field of Generative AI (e.g., Generative Adversarial Networks – GANs) has made significant advancements in recent times.  These advancements have led Generative AI to improve its capacity to produce highly realistic content such as artwork and images~\cite{zhang_styleswin_2022}. Hence, there is widespread utilization of Generative AI in both research and industry, leading to significant concerns regarding the protection of Intellectual Property Rights (IPR).
%
%
There is a surge of GAN-based image generative model usage for creating content that can infringe upon existing copyrights, spoof security systems, and undermine the rights of content owners through the spread of misinformation, affecting personal reputations. Consequently, AI-Generated images are facing lawsuits over alleged copyright violations. One of the notable examples of a legal dispute involving AI-Generated images includes the lawsuit filed by Getty Images and a group of artists against AI art generators.\footnote{https://www.businessinsider.com/ai-art-artists-getty-images-lawsuits-stable-diffusion-2023-1?op=1}
Additionally, producing state-of-the-art generative models requires substantial computational resources and large datasets. These issues have generated growing interest in research on protecting and verifying the ownership of the GAN models and training sets.

Previous copyright protection approaches for images and GANs include attacks (e.g., adversarial noise), watermarking, and attribution techniques. However, the literature focuses mainly on Deep Neural Networks (DNNs) and has a limited focus on developing copyright protection frameworks of GANs~\cite{chen_copy_2022}. 

Our research aims to fill this gap by conducting a comprehensive analysis of the effectiveness of the existing copyright protection approaches for GANs. The proposed framework assists in identifying the key factors that determine the effectiveness of copyright protection techniques through the analysis of various GAN models. In order to accomplish this goal, we explore and answer the following three research questions.

    \begin{itemize} [itemsep=0pt,partopsep=0pt, leftmargin=*]
        
        \item \textbf{RQ1:} Can current adversarial attacks  effectively prevent copyright violations of input images? 
        
        \item \textbf{RQ2:} 
        %
        %
        Can the IPR of GANs and their training sets be protected and verified through watermarking?
        %
        
       \item  \textbf{RQ3:} Can current attribution methods effectively support IPR protection and trace toxic generative AI model accountability through source attribution of output images?
    \end{itemize}

\begin{figure}[tp]
\centering
\includegraphics[width=0.75\linewidth]{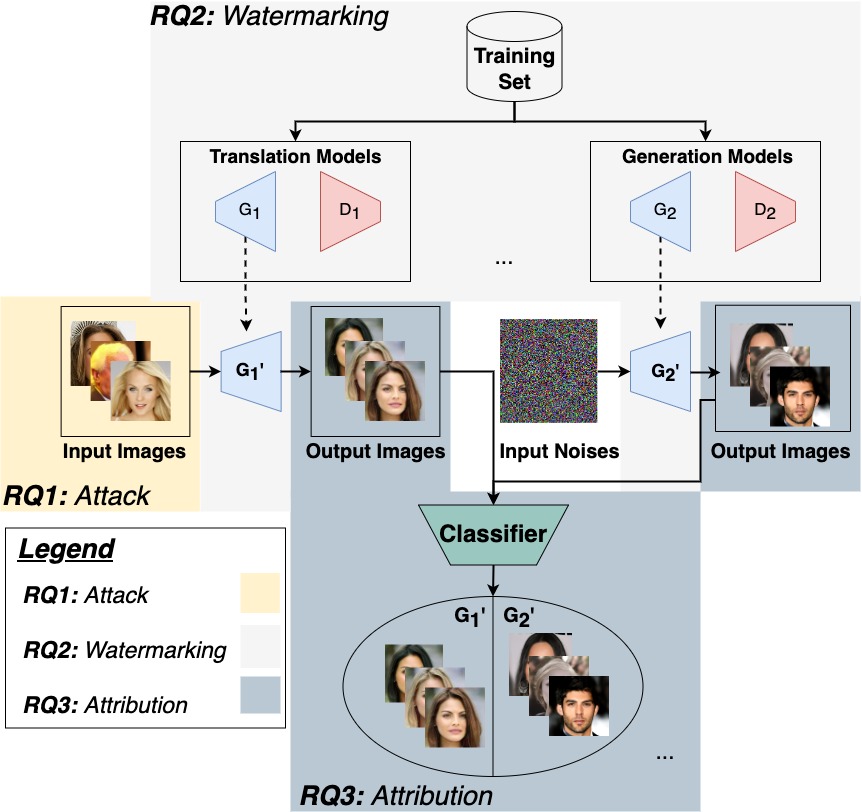}
\vspace{-0.2cm}
\caption{The proposed GAN and image copyright protection and accountability approach evaluation framework.}
\label{fig:flow}
\vspace{-0.2cm}
\end{figure}

\section{Experimental Setup}

This section provides the details of the experimental setup of the proposed work. Figure~\ref{fig:flow} provides a schematic representation of the framework utilized for evaluating the IPR protection methods on GANs.
%
%
To evaluate the effectiveness of different IPR approaches using the proposed evaluation framework, we utilize a range of datasets, including Horse2Zebra~\cite{zhu_unpaired_2017} for unconditional translation models and CelebA~\cite{liu2015faceattributes} and CelebAHQ~\cite{CelebAMask-HQ} for conditional translation and generation models. One of the main objectives of our work is to benchmark copyright protection techniques on GANs and their associated input sets (images) and outputs.  We performed this analysis on the following list of GAN architectures --- \textit{(i)} \textbf{Generation Models:} StyleGANv2~\cite{Karras2019stylegan2}, StyleGAN3~\cite{Karras2021}, StyleSwin~\cite{zhang_styleswin_2022}, DDGAN~\cite{xiao2022tackling}; \textit{(ii)} {\textbf{{Translation Models:}}
\emph{(a) Unconditional:} CycleGAN~\cite{zhu_unpaired_2017}, AttentionGAN-v2 (AttGAN)~\cite{tang2021attentiongan}, CUT~\cite{park2020cut}; \emph{(b) Conditional:} StarGAN~\cite{choi2018stargan}, StarGAN-v2~\cite{choi2020starganv2}.

\begin{table}[tp]
\normalfont
\centering
\caption{Performance of imperceptible Adversarial Attack on Different GANs. We followed the experimental configurations utilized in~\cite{ruiz2020disrupting} and set the regularization coefficient ($\epsilon$) to $0.05$ during all the experiments to enable a fair comparison. FID: Fréchet Inception Distance ~\cite{heusel_gans_2017}.}
\label{tab:table2}
 \resizebox{0.8\columnwidth}{!}{
\begin{tabular}{|c|c|c|c|c|}
\hline
\textbf{Model} & \textbf{Dataset} & \textbf{L1} & \textbf{L2} & \textbf{FID} \\ \hline
CUT & \multirow{3}{*}{Horse2Zebra} & 0.5229 & 0.4278 & 37.2135 \\ \cline{1-1} \cline{3-5} 
AttentionGAN &  & 0.5590 & 0.6153 & 227.4701 \\ \cline{1-1} \cline{3-5} 
CycleGAN &  & 0.4512 & 0.3084 & 74.4684 \\ \hline
StarGANv2 & \multirow{2}{*}{CelabAHQ $256 \times 256$} & 0.3553 & 0.2901 & 20.1240 \\ \cline{1-1} \cline{3-5} 
StarGAN &  & 1.0459 & 1.3470 & 251.7915 \\ \hline
StarGAN & CelabA $128 \times 128$ & 0.9288 & 1.1167 & 166.7310 \\ \hline
\end{tabular}
}
\end{table}

\section{RQ1: Adversarial Attacks}

    
        The copyright protection of input images is crucial in defending against the malicious modifications of images utilizing GANs. 
        %
        %
        One promising approach to counter fake image generation is to conduct adversarial perturbations on input images.
        
        \noindent \textbf{Related work.} There has been extensive discussion on the application of invisible adversarial perturbations to input images to disrupt the illicit use of generative models for fake image generation. Previous research has successfully implemented attacks on both unconditional and conditional image-translation models. Consequently, adversarial attacks have become a promising tool for ensuring that copyrighted images are not subjected to unauthorized modifications by GANs.
        Hence, we examine whether current adversarial watermarks can lead to satisfactory copyright protection against the illicit use of the input images. 
        %
        %
        For this task, we employ adversarial watermarks generated via the utilization of the Projected Gradient Descent (PGD) algorithm trained with model outputs, as proposed in Ruiz \etal's work~\cite{ruiz2020disrupting}. This technique is widely established in the field of image watermarking and has been frequently employed as one of the benchmarked approaches in many recent attacks~\cite{huang_cmua-watermark_2022}.

    \noindent \textbf{Experimental results and discussion.}
        We employ adversarial watermarks to generate adversarial examples to attack models that have not been assessed qualitatively or quantitatively. We measured the disturbance levels by calculating the L1 and L2 norms and FID~\cite{heusel_gans_2017} on the original output of the generative network and the output generated with adversarial input examples. 

        \begin{figure}[tp]
        \centering
        \includegraphics[width=0.94\linewidth]{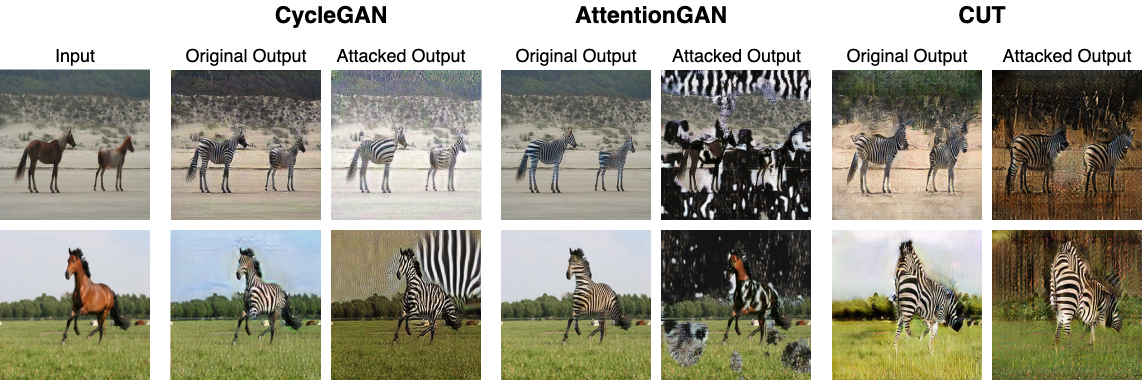}
    \vspace{-2mm}
        \caption{Original and attacked output for unconditional translation GANs using the Horse2Zebra dataset.}
        \label{fig:attack_h2z}
        
        \includegraphics[width=0.97\linewidth]{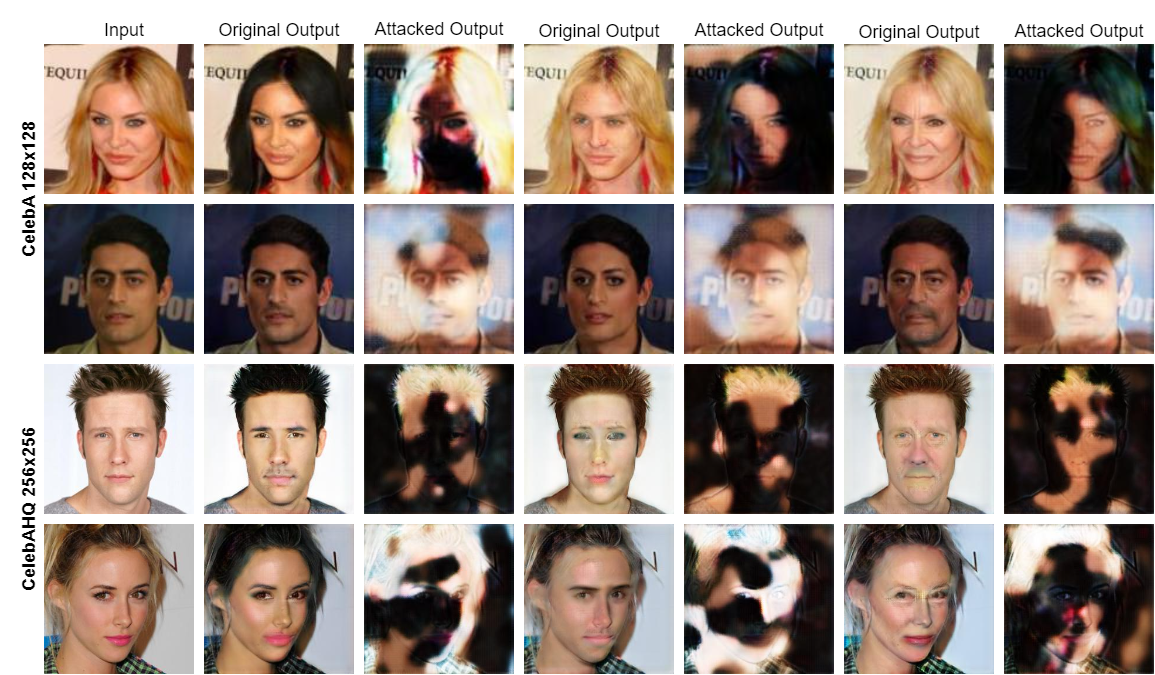}
            \vspace{-2mm}
        \caption{Original and attacked output for StarGAN with different image qualities.}
        \vspace{1mm}
        \label{fig:attack_starGAN_quality}

        \includegraphics[width=0.94
        \linewidth]{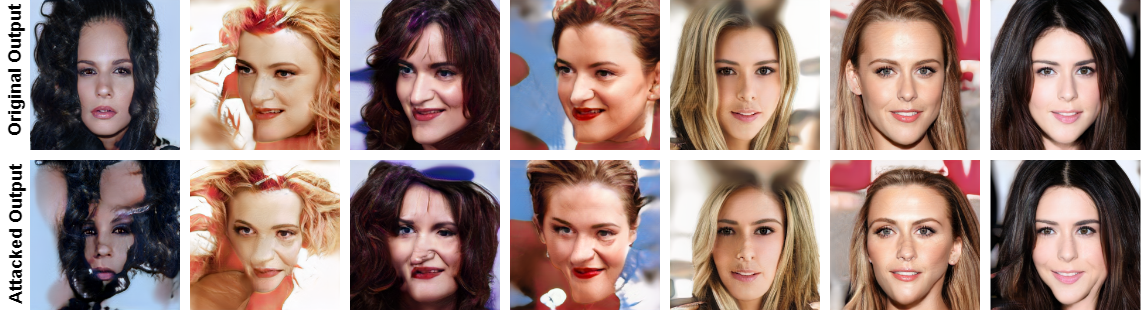}
        \vspace{-2mm}
        \caption{Original and attacked output for StarGANv2 using the CelebAHQ dataset.}
        \label{fig:attack_starganv2}
        \end{figure}

        As shown in Table~\ref{tab:table2}, the adversarial attack causes significant disruptions in outputs. However, there is a complex interplay of factors  that has not been investigated before. Consequently, to enable a more in-depth understanding of the similarity between original and generated images, we use FID~\cite{heusel_gans_2017} together with L1 and L2 norm calculations. Through the experiments, adversarial attacks on images with high resolution deliver a better attack performance compared to the attacks on lower resolution images (Figure~\ref{fig:attack_starGAN_quality}). Additionally, the models with better performance (AttentionGAN and CycleGAN) will lead to more perceivable disruptions (Figure~\ref{fig:attack_h2z}).

        Figure~\ref{fig:attack_h2z} also illustrates that AttentionGAN suffered from the highest level of distortion, with the background of the images being significantly affected.
        We conjecture that this is related to the attention mechanism of AttentionGAN
        and it is due to the presence of noise perturbing the model's ability to distinguish the objects of interest from the noisy background.
        
        %
        The output generated by StarGANv2 gathered a significant interest due to the low FID value.
        Figure~\ref{fig:attack_starganv2} demonstrates that the distortion caused by the adversarial attack is less significant when the face is clearly visible. However, in both cases, the style of the source inputs is preserved. Hence, we conclude that the disruption is more targeted to the source characteristic preservation~\cite{choi2020starganv2}, while less effective in preventing style reconstruction. However, further investigation into the relevant loss is necessary to confirm this hypothesis. Furthermore, it shows the limitation of FID in assessing the distortions when the style of images resembles the source image.

\vspace{1mm}     
\begin{center}
\resizebox{0.99\columnwidth}{!}{

\begin{tcolorbox}
\textbf{RQ1. Takeaways:}
\begin{itemize} [leftmargin=*]

\item Recent developments in invisible adversarial attack-based watermarks on input images have facilitated the effective defence against the unauthorized use of images.

\item New metrics are needed to comprehensively assess the effectiveness of attacks, as current metrics exhibit limitations in accurately reporting distortions.

\item Adversarial attack watermarks can be less effective in applications focusing on style transfer compared to source characteristic preservation in image-to-image translation.
\end{itemize}
\end{tcolorbox}}
\end{center}

\section{RQ2: Watermarks}

        Watermarking is crucial for the copyright protection of GAN models and training sets, given the cost of efforts and resources required to develop advanced GAN architectures, gather comprehensive datasets and train those models rigorously. The goal of watermarking GANs is to embed watermarks within a model and later retrieve them from the generated content. These watermarks can be later used as evidence in the event of unauthorized utilization of copyrighted models and datasets.
        
        \noindent \textbf{Related work.} There are two potential approaches that could be employed to achieve watermarking. One extensively researched method uses DNNs to create a watermark network. This approach involves training a watermark encoder to embed the watermark while training the model and a decoder to retrieve the watermark from the generated content. Yu \etal~\cite{yu_artificial_2021} approached this concept by training an encoder and a decoder first, and then superimposing the fingerprints on training sets for GAN training. The watermark will be attached to the GAN, so the fingerprints can be detected in the generated images. Fei \etal~\cite{fei_supervised_2022} extended this approach by integrating a decoder and an encoder into GAN training. For our analysis, we chose Yu \etal's approach~\cite{yu_artificial_2021} for our evaluation, as it provides IPR protection strategies on both training sets and models. 

        Another branch proposed by Ong \etal~\cite{ong_protecting_2021} is to embed the ownership information into the model as a regularization term so that a trigger input into the suspect model will generate an output with the watermark. It essentially creates a backdoor for the copyright owners to employ a trigger image to verify whether a black-box API is using their models.
        
    \noindent \textbf{Experimental results and discussion.} We utilized Yu's method~\cite{yu_artificial_2021} as a representative CNN-based watermarking technique and the Ong~\cite{ong_protecting_2021} to create a model watermark backdoor. 
    %
    %
    During the implementation of the watermarks, we evaluated their performance across multiple epochs to determine their optimal configurations.
    
        \begin{figure}[tp]
        \centering
        \includegraphics[width=1\linewidth]{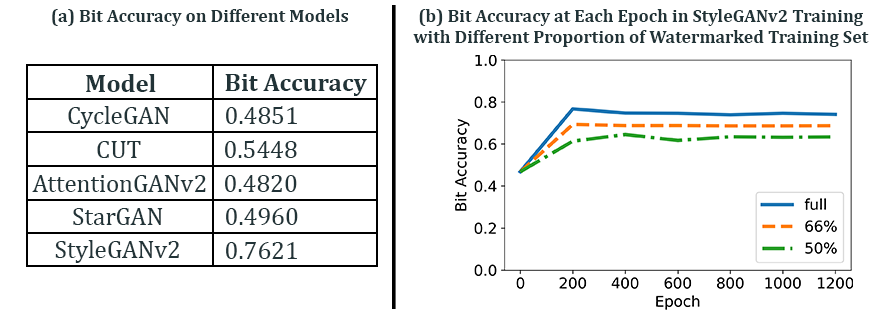}
        \vspace{-0.5cm}
        \caption{Bit accuracy of watermarking detection based on Yu \etal \cite{yu_artificial_2021}. (a) Bit accuracy of watermark retrieval on Horse2Zebra (CycleGAN, CUT, and AttentionGAN-v2) and CelebA (StarGAN and StyleGANv2). (b) Bit accuracy at different epochs of GAN training.}
    
        \label{fig: watermark_pp}
        \vspace{0.2cm}
        \includegraphics[width=0.9\linewidth]{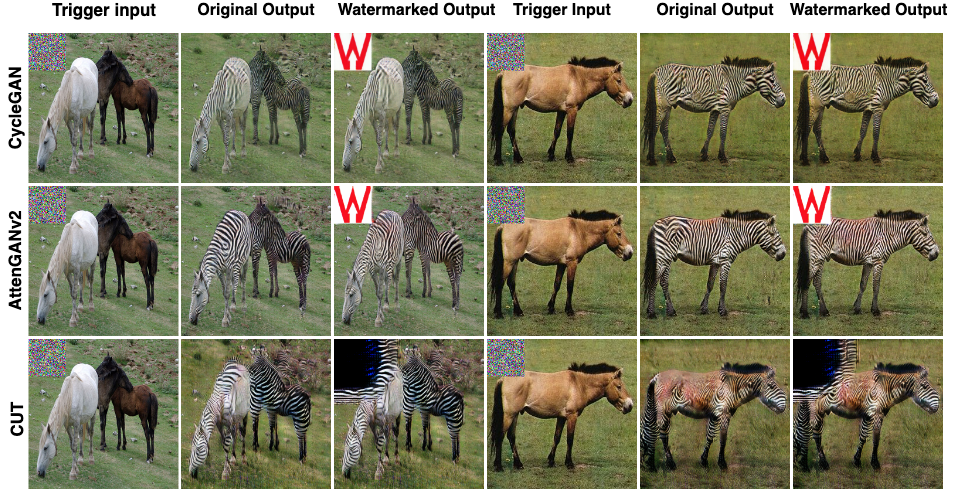}
        \vspace{-0.2cm}
        \caption{Original, trigger, and watermarked images of watermark injection method on unconditional translation models.}
        \label{fig: ipr_concat}
        \end{figure}

        Figure~\ref{fig: watermark_pp} shows the result of dataset watermarking on different models. We used the watermarked datasets to train models and conducted watermark detection. We find that it is challenging to watermark translation models, whereas the performance of the generation model in terms of bit accuracy is improved. We assume that it is due to the generation model's higher reflection of the original training sets in comparison to translation models. 
        Moreover, watermarks tend to be learned during the early training stages of generation GAN model training, regardless of the proportion of the watermarked images in the training set. This feature is important since it indicates that watermarking models can be a less demanding task than expected, given that current models can be easily watermarked by being retrained on the watermarked datasets.
        
        We also utilized the regularization approach on translation models and used trigger inputs to test the performance of all three unconditional translation models. Overall, the results generated from the regularization approach show that the trigger image can successfully produce a watermarked output without affecting the performance of the models. However, the CUT model fails to reproduce the watermark image. We assume that it is due to its patch-wise training mechanism. This issue raises the importance of an in-depth investigation into the limitations of current watermarking methods to achieve IPR protection.

        \begin{tcolorbox}
        \textbf{RQ2. Takeaways:}
        \begin{itemize} [leftmargin=*]
        
        \item Recent developments in watermarks have led to improved protection of GAN models. However,  IPR protection for training sets remains a prevalent concern.
        
        \item Watermark detection capabilities of generation models trained on watermarked training sets are better than those of translation models.
        
        \end{itemize}
        \end{tcolorbox}

\section{RQ3: Attribution}


        The attribution of GAN architectures 
        %
        %
        also plays a crucial role in copyright protection.
        It allows IPR owners to differentiate between real images 
 and fake images, and their source models by attributing fake images based on the architectural characteristics of the underlying model. There are two broad categories of attribution methods: spectral and contrastive.
    
       \noindent \textbf{Related work.} The utilization of spectral analysis was proven to be effective in DeepFake detection and model attribution. Consequently, detectors based on discrete cosine transformation (DCT)~\cite{frank_leveraging_2020} were proposed and validated. Contrastive attribution is another approach used to determine the modifications a GAN model has made to a generated image. This is achieved by conducting analyses on local feature differences between original and generated images.  Yang~\etal~\cite{yang2022deepfake} discussed patch-wise contrastive learning to generate predictions based on the model fingerprints.
        
        \begin{table}[tp]
    \centering
    \caption{The confusion matrix for attribution  of ``Translation GANs'' using Horse2Zebra with spectral analysis (Rows: True Values, Columns: Predicted Values).}
    \label{tab:generation}
    \resizebox{0.6\columnwidth}{!}{
    \begin{tabular}{|l|l|l|l|l|}
    \hline
     & \textit{\textbf{AttGAN}} & \textit{\textbf{CUT}} & \textit{\textbf{CycleGAN}} & \textit{\textbf{Real}} \\ \hline
    \textit{\textbf{AttGAN}} & \textbf{209} & 3 & 0 & 25 \\ \hline
    \textit{\textbf{CUT}} & 5 & \textbf{221} & 7 & 4 \\ \hline
    \textit{\textbf{CycleGAN}} & 0 & 2 & \textbf{233} & 2 \\ \hline
    \textit{\textbf{Real}} & 19 & 2 & 0 & \textbf{216} \\ \hline
    \textbf{Precision} & \textbf{0.8970} & \textbf{0.9693} & \textbf{0.9708} & \textbf{0.8745} \\ \hline
    \end{tabular}
    }
    \vspace{2mm}
    \caption{The confusion matrix for attribution for ``Generation GANs'' on 5000 images generated using CelebAHQ with spectral analysis (Rows: true values, Columns: predicted values).}
    \label{tab:translation}
    \resizebox{0.9\columnwidth}{!}{
    \begin{tabular}{|l|l|l|l|l|l|l|}
    \hline
    \textbf{} & \textit{\textbf{DD-GAN}} & \textit{\textbf{StarGAN}} & \textit{\textbf{StarGANv2}} & \textit{\textbf{StyleGAv2}} & \textit{\textbf{StyleSwin}} & \textit{\textbf{Real}} \\ \hline
    \textit{\textbf{DD-GAN}} & \textbf{934} & 13 & 0 & 33 & 0 & 20 \\ \hline
    \textit{\textbf{StarGAN}} & 4 & \textbf{996} & 0 & 0 & 0 & 0 \\ \hline
    \textit{\textbf{StarGANv2}} & 0 & 1 & \textbf{992} & 0 & 3 & 4 \\ \hline
    \textit{\textbf{StyleGANv2}} & 3 & 1 & 0 & \textbf{992} & 2 & 2 \\ \hline
    \textit{\textbf{StyleSwin}} & 0 & 0 & 12 & 10 & \textbf{962} & 16 \\ \hline
    \textit{\textbf{Real}} & 36 & 6 & 43 & 61 & 77 & \textbf{777} \\ \hline
    \textbf{Precision} & \textbf{0.956} & \textbf{0.9794} & \textbf{0.9475} & \textbf{0.9051} & \textbf{0.9215} & \textbf{0.9487} \\ \hline
    \end{tabular}
    }
    \end{table}
    
    \noindent \textbf{Experimental results and discussion.}
    %
    Following the approach and extending the scope outlined in Frank \etal's 
    work~\cite{frank_leveraging_2020}, we generated the confusion matrix for attribution spectral analysis. 
    For this analysis, we generated 237 fake images from the Horse2Zebra dataset for each translation model, and generated 5,000 fake images from the CelebAHQ for each generative model.

    The results indicate that the accuracy generated by spectral analysis is high when trained on different datasets. The results are as expected, given that discrete cosine transformation produces distinct fingerprints for each model (Figure~\ref{fig: spectrum}). 
    %
    %
    The empirical results further confirm that the models with more capabilities exhibit better performance in generating fake images that are more realistic (refer to Table~\ref{tab:translation}).

    We are conducting an ongoing experiment on the contrastive method and attribution evasion proposed by Dong \etal~\cite {dong_think_2022}. However, the evasion pipeline requires modifications to the network, which contradicts the requirement of the watermark to remain  intact during IPR protection. Hence,  the authors are not motivated to incorporate spectral loss into their model.

    \begin{tcolorbox}
    \textbf{RQ3. Takeaways:}
    \begin{itemize} [leftmargin=*]
    
    \item 
    %
    %
    Classifiers that utilize spectral domain fingerprints can achieve IPR protection of the model through the attribution of the output images.
    \item Spectral features can be harder to be traced when the models generate more realistic outputs.
    \end{itemize}
    \end{tcolorbox}
\section{Conclusion}

This study conducted a preliminary evaluation of the state-of-the-art IPR protection methods for GANs and their training sets (images).     
    %
    %
    %
    The experimental results demonstrate that the recently developed IPR protection techniques such as adversarial attacks, watermarking, and attribution are largely effective for copyright protection and accountability tracing of GAN models. 
    %
    %
    However, our findings reveal that the current techniques for protecting the IPR of the training sets (the original copyrighted images) are inadequate.
    %
    %
    We also highlight the importance of further analysis of recent generative AI models that are capable of generating realistic images utilizing the unauthorized styles and works of IPR owners.     Furthermore, this paper underscores the importance of developing a unified pipeline to assess the performance of adversarial attack watermarks and measure the IPR protection levels and accountability tracing of generative AI. Our ongoing research aims to integrate 
    %
    %
    findings from various domains of IPR protection across a plethora of generative AI models to develop a comprehensive set of metrics. The findings derived from this preliminary study will inform lawmakers and practitioners of ethical generative AI adoption. 
    
    \begin{figure}[tp]
    \includegraphics[width=1\linewidth]{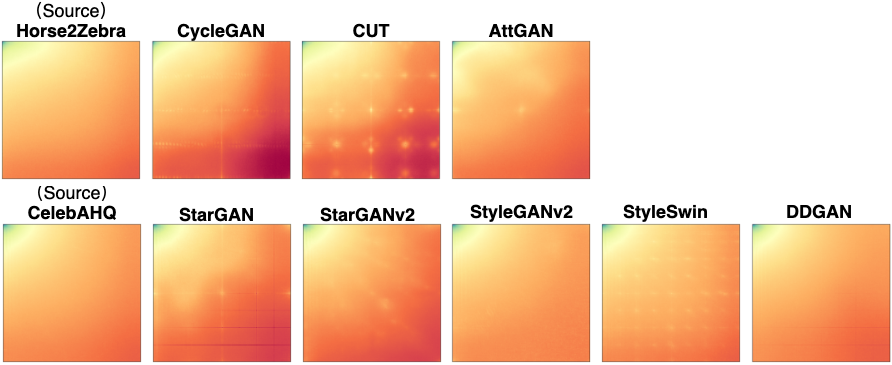}
    \vspace{-4mm}
    \caption{Spectral Imprints Produced by GANs using DCT.}
    \label{fig: spectrum}
       \vspace{-3mm}
    \end{figure}



\end{document}